\def\year{2022}\relax
\documentclass[letterpaper]{article} 
\usepackage{aaai22}  
\usepackage{times}  
\usepackage{helvet}  
\usepackage{courier}  
\usepackage[hyphens]{url}  
\usepackage{graphicx} 
\usepackage{natbib}  
\usepackage{caption} 
\DeclareCaptionStyle{ruled}{labelfont=normalfont,labelsep=colon,strut=off} 
\usepackage{algorithm}
\usepackage{algorithmic}

\usepackage{amsmath}
\usepackage{amssymb}
\usepackage{newfloat}
\usepackage{listings}
\floatstyle{ruled}
\newfloat{listing}{tb}{lst}{}
\floatname{listing}{Listing}
\title{AdaptivePose: Human Parts as Adaptive Points}
\author {
    Yabo Xiao,\textsuperscript{\rm 1}
    Xiaojuan Wang, \textsuperscript{\rm 1,}\footnote{Corresponding author.} 
    Dongdong Yu, \textsuperscript{\rm 2}
    Guoli Wang, \textsuperscript{\rm 3}
    Qian Zhang, \textsuperscript{\rm 4}
    Mingshu He \textsuperscript{\rm 1}
}
\affiliations {
    \textsuperscript{\rm 1} Beijing University of Posts and Telecommunications\\
    \textsuperscript{\rm 2} ByteDance Inc.\\
    \textsuperscript{\rm 3} Tsinghua University\\
    \textsuperscript{\rm 4} Horizon Robotics\\
    \{xiaoyabo, wj2718, hemingshu\}@bupt.edu.cn, yudongdong@bytedance.com, wangguoli1990@mail.tsinghua.edu.cn, qian01.zhang@horizon.ai
}

\begin{document}

\maketitle

\begin{abstract}

  Multi-person pose estimation methods generally follow top-down and bottom-up paradigms, both of which can be considered as two-stage approaches thus leading to the high computation cost and low efficiency. Towards a compact and efficient pipeline for multi-person pose estimation task, in this paper, we propose to represent the human parts as points and present a novel body representation, which leverages an adaptive point set including the human center and seven human-part related points to represent the human instance in a more fine-grained manner. The novel representation is more capable of capturing the various pose deformation and adaptively factorizes the long-range center-to-joint displacement thus delivers a single-stage differentiable network to more precisely regress multi-person pose, termed as AdaptivePose. For inference, our proposed network eliminates the grouping as well as refinements and only needs a single-step disentangling process to form multi-person pose.~Without any bells and whistles, we achieve the best speed-accuracy trade-offs of 67.4\% AP / 29.4 fps with DLA-34 and 71.3\% AP / 9.1 fps with HRNet-W48 on COCO test-dev dataset. 
  
\end{abstract}

\section{Introduction}

With the prevalence of deep learning technique, Pose estimation \cite{YaboXiao2020SPCNetSpatialPA, AlejandroNewell2016StackedHN} has drawn much attention in computer vision area. It's an essential step for many high-level vision tasks such as activity understanding~\cite{shi2019two,li2019actional}, pose tracking~\cite{xiao2018simple} and so on. Most existing methods for multi-person pose estimation can be summarized in two ways including the top-down methods~\cite{chen2018cascaded,su2019multi,sun2019deep, fang2017rmpe} and the bottom-up methods~\cite{cao2017realtime,kreiss2019pifpaf, newell2017associative, papandreou2018personlab,cheng2020higherhrnet}. The top-down methods crop and resize the region of detected person firstly and then locate the single-person keypoints in each cropped region. These methods may have the following drawbacks: 1) The performance of joint detection is strongly tied to the quality of the human bounding boxes. 2) Detection-first pattern leads to high memory cost and low efﬁciency and is not feasible for applications. The bottom-up methods firstly locate the keypoints for the all persons simultaneously and then group them for each person. Although bottom-up methods generally run faster than top-down methods, the grouping process is still computationally complexity and redundant, and always involves many tricks to refine the final results. 

\begin{figure}
\begin{center}
\includegraphics[height=0.56\columnwidth,width=1.0\columnwidth]{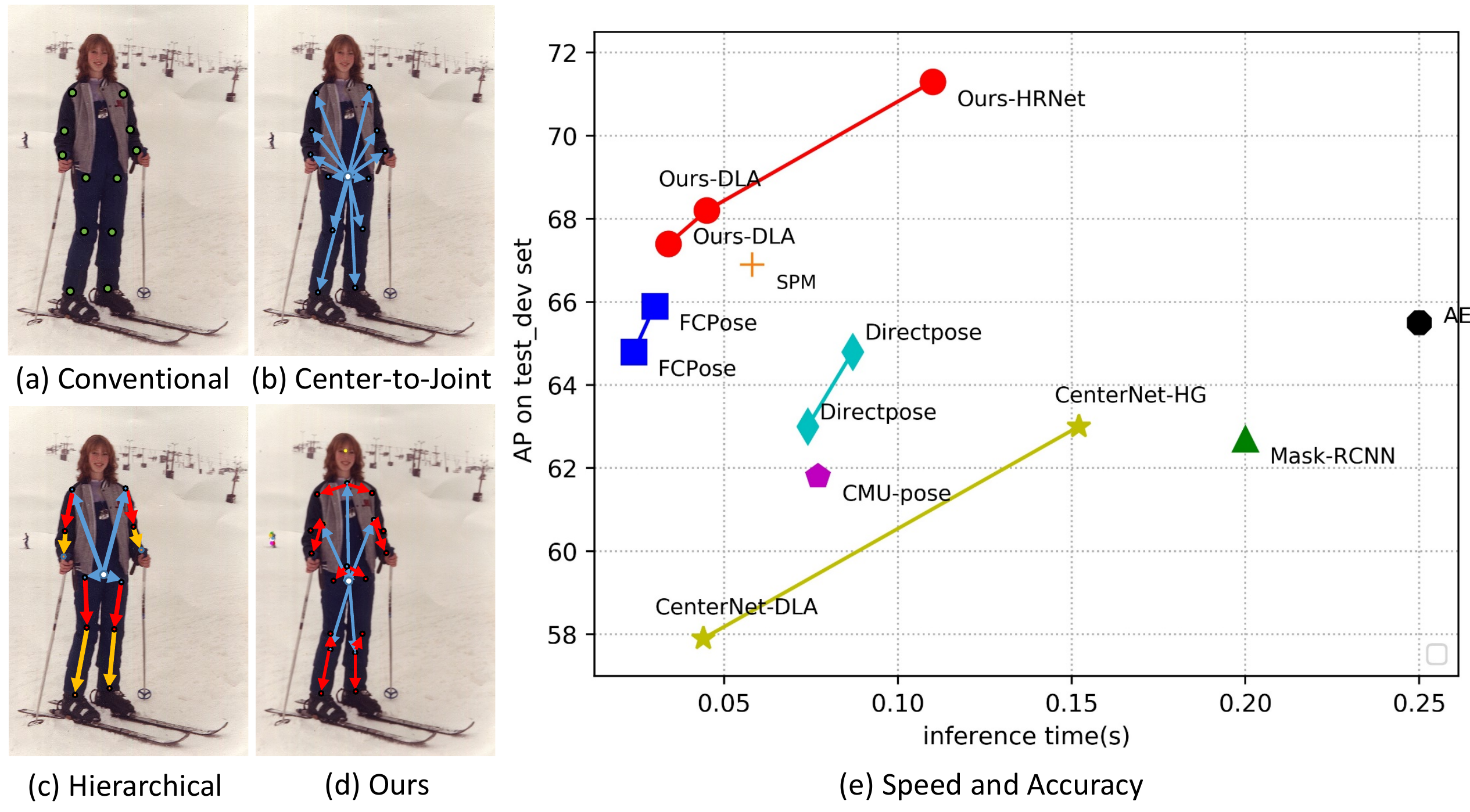}
\end{center}
\caption{(a) Conventional body representation generally used in top-down methods such as Mask-RCNN \cite{he2017mask} and Rmpe \cite{fang2017rmpe} as well as bottom-up methods such as CMU-pose \cite{cao2017realtime} and AE \cite{newell2017associative}. (b) Center-to-joint body representation proposed by CenterNet \cite{zhou2019objects}. (c) Hierarchical structured body representation introduced by SPM \cite{nie2019single}. (d) An adaptive point set representation proposed by our method. (e) Inference time~(s) versus precision~(AP). Our method achieves the best speed-accuracy trade-offs compared with the representative bottom-up and single-stage methods.}

\label{fig:image1}
\end{figure}

Aforementioned two-stage methods generally use the conventional representation that models the human pose via absolute keypoint position as shown in Figure \ref{fig:image1}(a), which separates the associations between the human instance and keypoints thus requires an extra stage to model the relationship. Recent research works preliminarily explore the representation methods to model the relation between human instance and corresponding keypoints while suffer some obstacles thus leading to the limited performance. For example, as shown in Figure \ref{fig:image1}(b), CenterNet \cite{zhou2019objects} represents the human instance via center point and leverages center-to-joint offsets to form human pose but achieves the compromised performance since various pose deformation and the center has fixed receptive field thus hard to deal with long-range center-to-joint offsets. As shown in Figure \ref{fig:image1}(c), SPM \cite{nie2019single} also represents the instance via root joint and further presents a fixed hierarchical tree-structure and divides the root joint and keypoints into four hierarchies based on articulated kinematics. It factorizes the long-range offsets into accumulative short ones while suffers the dilemma of accumulated error propagated along the skeleton.

To tackle with the aforementioned problems, In this work, we propose to represent the human parts as adaptive points and use an adaptive point set including the human center and seven human-part related points to fit diverse human instances. The human pose is formed in a body (center)-to-part (adaptive points)-to-joint manner, as shown in Figure \ref{fig:image1}(d). Compared with previous representations, the superiorities of our representation mainly involve in two aspects: 1) This fine-grained point set representation is more capable of capturing the various extent of deformation for human body compared with center representation. 2) It adaptively factorizes the long-range displacement into shorter ones while avoids the accumulated error propagated along the skeleton since the adaptive human-part related points are automatically learned by neural network.

Based on the adaptive point set representation, we propose an efficient end-to-end differentiable network, termed as AdaptivePose, which mainly consists of three novel components. First, we present a Part Perception Module to perceive the human parts by dynamically predicting seven adaptive human-part related points for each human instance. Second, in contrast to using the feature with fixed receptive field to predict the center of various bodies, Enhanced Center-aware Branch is introduced to conduct the receptive field adaptation and capture the various pose deformation by aggregating the features of adaptive human-part related points to perceive the center more precisely. Third, we propose a Two-hop Regression Branch where the adaptive human-part related points act as one-hop nodes to dynamically factorize long-range center-to-joint offsets. During inference, we only need a single-step decode process via combining the center positions and center-to-joint offsets to form human pose without any refinements and tricks.

\begin{figure}
\begin{center}
\includegraphics[height=0.478\columnwidth,width=1.0\columnwidth]{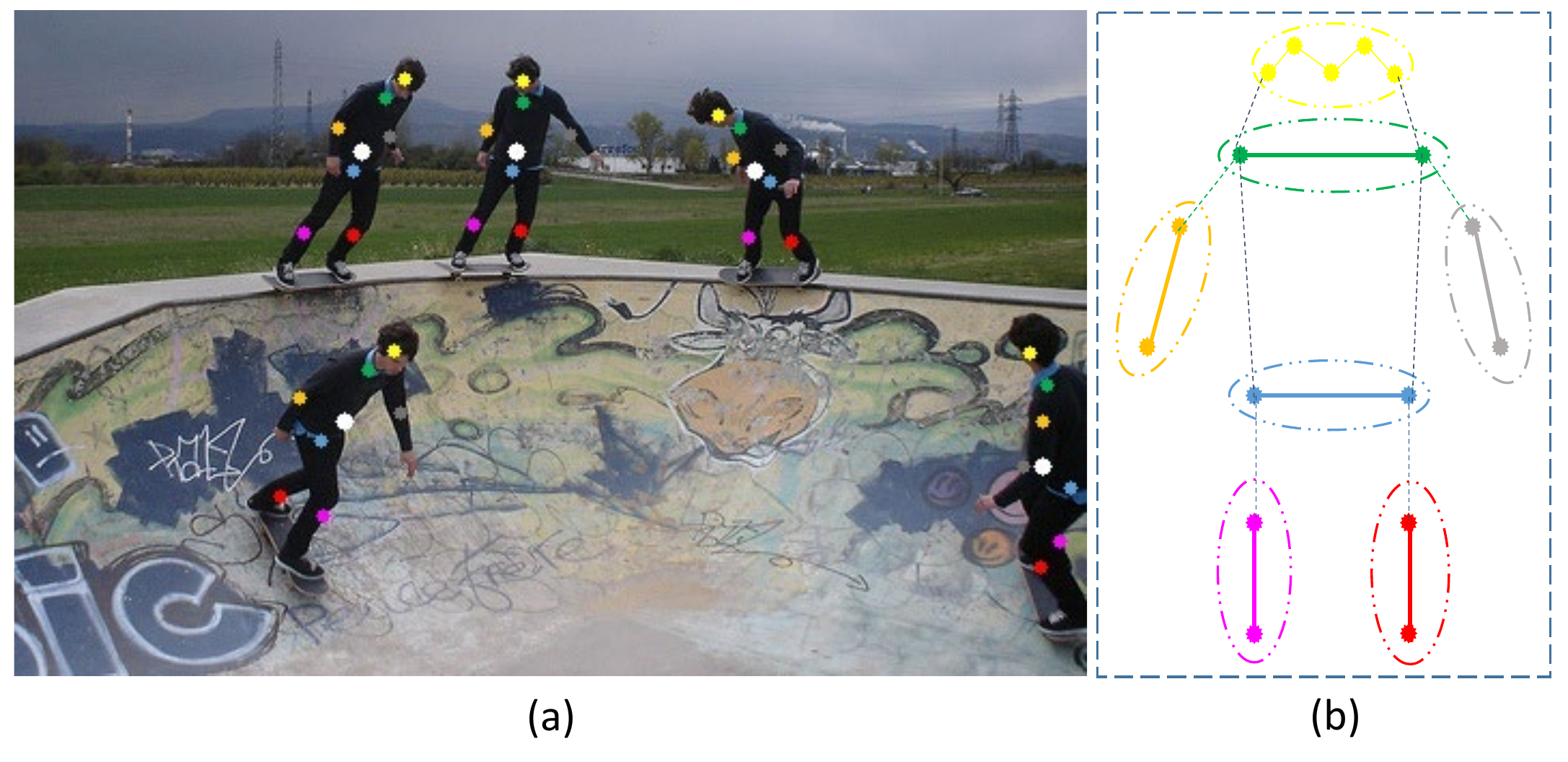}
\end{center}
\caption{(a) The visualization of adaptive point set. White points indicate the human center and others refer to human-part related points visualized by colorful points corresponding the part with same color in Figure (b). (b) Divided human parts according to the degrees of freedom and extent of deformation. We leverage an adaptive point set conditioned on each human instance to represent the human in a fine-grained way.}
\label{fig:image2}
\end{figure} 


The main contributions can be summarized as follows:
\begin{itemize}

\item We propose to represent human parts as points and further leverage an adaptive point set to represent the human instances. To our best knowledge, we are the first to present the fine-gained and adaptive body representation, which is more capable of capturing the various pose deformation and adaptively factorizes the long-range center-to-joint offsets.



\item Based on the novel representation, we propose a compact single-stage differentiable network, termed as AdaptivePose. Specifically, we introduce a novel Part Perception Module to perceive the human parts by regressing seven human-part related points. By using human-part related points, we further propose an Enhanced Center-aware Branch to more precisely perceive the human center and a Two-hop Regression Branch to effectively factorize the long-range center-to-joint offsets.

\item Our method significantly simplifies the pipeline of multi-person pose estimation and achieves the best speed-accuracy trade-offs of 67.4\% AP / 29.4 fps , 68.2\% AP / 22.2 fps with DLA-34, and 71.3\% AP / 9.1 fps with HRNet-W48 on COCO test-dev set without any refinements and post-process.

\end{itemize}

\begin{figure*}
\begin{center}
\includegraphics[height=0.85\columnwidth,width=2.1\columnwidth]{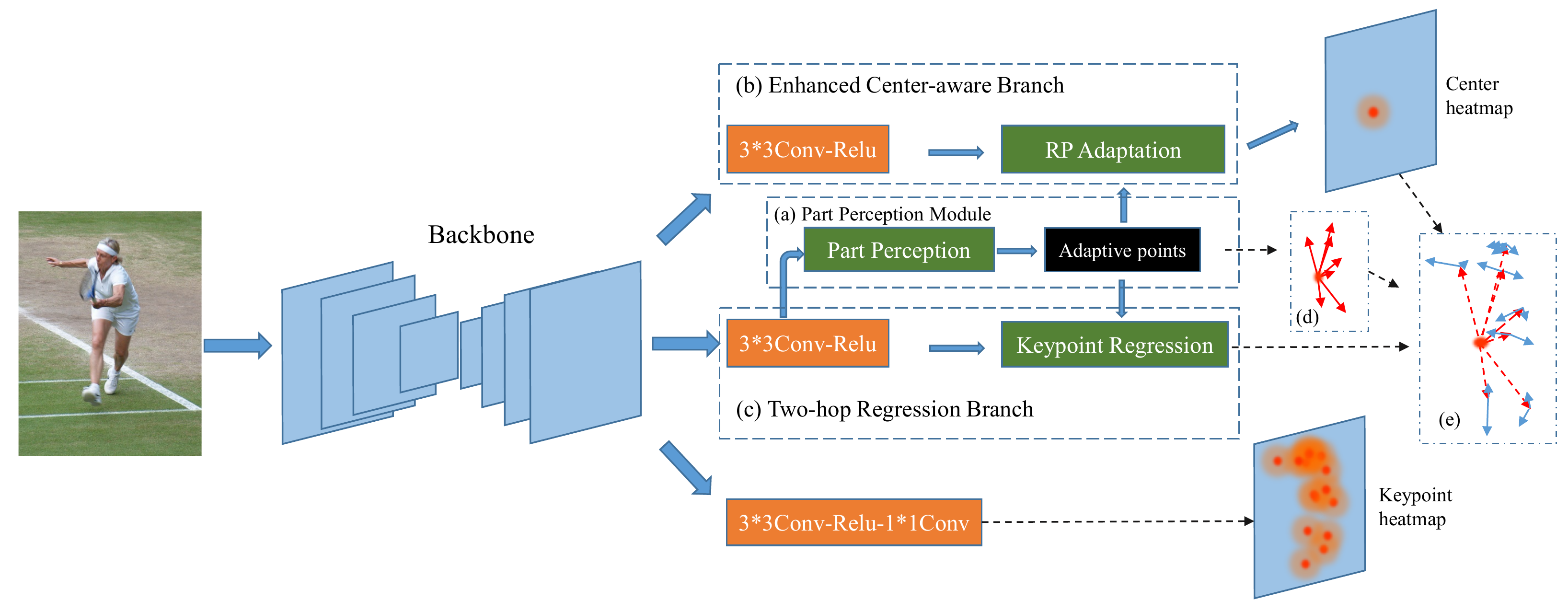}
\end{center}
\caption{ Overview of AdaptivePose.~(a) The structure of Part Perception Module, Adaptive points indicate seven human-part related points.~(b) The structure of Enhanced Center-aware Branch, RP Adaptation refers to receptive field adaptation.~(c) The diagram of Two-hop Regression Branch.~(d) The red arrows are one-hop offsets that dynamically locate the adaptive human-part related points.~(e) The blue arrows indicate second-hop offsets for localizing the human joints.}
\label{fig:image3}
\end{figure*}


\section{Related work}
In this section, we review three parts related to our method including top-down methods, bottom-up methods and point-based representation methods.  

{\bf Top-down Methods.} Given an arbitrary RGB image, top-down methods firstly detect the location of human instance and then locate their keypoints individually. Concretely, the region of each human body would be cropped and resized to the unified size so that it has superior performance. Top-down methods mainly focus on the design of the network to extract better feature representation. HRNet~\cite{sun2019deep} maintains high-resolution representations and repeatedly fuses multi-resolution representations through the whole process to generate  reliable high-resolution representations. \cite{su2019multi} proposes a Channel Shuffle Module and Spatial, Channel-wise Attention Residual Bottleneck~(SCARB) to drive the cross-channel information flow. However, due to inefficiency caused by the detection-first paradigm, top-down methods are often not feasible for the real-time systems with strict latency constraints.

{\bf Bottom-up Methods.} In contrast to top-down methods, bottom-up methods localize keypoints of all human instances with various scales at first and then group them to the corresponding person. Bottom-up methods mainly concentrate on the effective grouping process. For example, CMU-pose~\cite{cao2017realtime} proposes a non parametric representation, named Part Affinity Fields~(PAFs), which encodes the location and orientation of limbs, to group the keypoints to individuals in the image. AE~\cite{newell2017associative} simultaneously outputs a keypoint heatmap and a tag heatmap for each body joint, and then assigns the keypoints with similar tags into individual. However, one case worth noting is that the grouping process serves as a post-process is still computationally complex and redundant. 


{\bf Point-based Representation.} The keypoint-based methods \cite{lee2020centermask,law2018cornernet,zhou2019objects,nie2019single,tian2020conditional} represent the instances by center or paired corners and have been applied in many tasks. They have drawn much attention as they are always simpler and more efficient than anchor-based representation \cite{cai2018cascade,ren2015faster,Lin_2017_CVPR,lin2017focal,huang2019mask,liu2018path}. CenterNet proposes to use keypoint estimation to find center and then regresses the other object properties such as size to predict bounding box. SPM \cite{nie2019single} represents the person via root joint and further presents a fixed hierarchical body representation to estimate human pose. Point-Set Anchors \cite{wei2020point} propose to leverage a set of  pre-defined points as pose anchor to provide more informative features for regression. DEKR \cite{geng2021bottom} leverages the center to model the human instance and use a multi-branch structure that adopts adaptive
convolutions to focus on each keypoint region for separate keypoint regression. In contrast to previous methods that use center or pre-defined pose anchor to model human instance, we propose to represent human instance via an adaptive point set including center and seven human-part related points as shown in Figure \ref{fig:image2}(a). The novel representation is able to capture the diverse deformation for human body and adaptively factorize long-range displacements.



\section{Method}

In this section, we firstly introduce our proposed body representation. Then, we elaborate on network architecture including each component. Finally, we report the training and inference details.               
 
\subsection{Body Representation}\label{subsection3.1}
In contrast to previous body representation methods, we present an adaptive point set representation that uses the center point and seven human-part related points to represent the human instance in a fine-grained manner. The proposed representation introduces the adaptive human-part related points, which are used to finely-grained capture the structured human pose with various deformation and adaptively factorize the long-range center-to-joint offsets into shorter ones while avoids the accumulated error propagated along the fixed articulated skeleton.

In particular, we manually divide the human body into seven parts (i.e., face, shoulder, left arm, right arm, hip, left leg and right leg) according to the inherent structure of human body, as shown in Figure~\ref{fig:image2}(b). Each divided human part is represented by an adaptive human-part related point, which is dynamically regressed from the human center. The process can be formulated as:

\begin{equation} \label{one-hop}
  \mathbf{C}_{inst} \rightarrow \left \{\mathbf{P}_{face}, \mathbf{P}_{sho}, \mathbf{P}_{la}, \mathbf{P}_{ra}, \mathbf{P}_{hip}, \mathbf{P}_{ll}, \mathbf{P}_{rl}\right\},
\end{equation}
where $\mathbf{C}_{inst}$ refers to the instance center, others indicate seven adaptive human-part related points corresponding to face, shoulder, left arm, right arm, hip, left leg and right leg. Human instance $inst$ is finely-grained represented by a point set $\left \{\mathbf{C}_{inst}, \mathbf{P}_{face}, \mathbf{P}_{sho}, \mathbf{P}_{la}, \mathbf{P}_{ra}, \mathbf{P}_{hip}, \mathbf{P}_{ll}, \mathbf{P}_{rl}\right\}$. For convenience, $\mathbf{P}_{part}$ is used to indicate the seven human-part related points $\mathbf{P}_{face}, \mathbf{P}_{sho}, \mathbf{P}_{la}, \mathbf{P}_{ra}, \mathbf{P}_{hip}, \mathbf{P}_{ll}, \mathbf{P}_{rl}$. Then we leverage human-part related points to locate the keypoints belong to corresponding parts as follows:

\begin{equation} \label{two-hop}
\mathbf{P}_{part} \rightarrow \mathbf{Joint}.
\end{equation}
The novel representation starts from instance-wise (body center) to part-wise (adaptive human-part related points), then to joint-wise (body keypoints) to form human pose.

This fine-grained representation delivers a single-stage solution thus we present to build a single-stage differentiable regression network to estimate multi-person pose, where Part Perception Module is proposed to predict seven human-part related points. By using the adaptive human-part related points, Enhanced Center-aware Branch is introduced to perceive the center of human with various pose deformation and scales. In parallel, Two-hop Regression Branch is presented to regress keypoints via center-to-part-to-joint manner.

\subsection{Single-stage Network}\label{subsection3.2}

{\noindent\bf Overall Architecture.} As shown in Figure \ref{fig:image3}, given an input image, we first extract the general semantic feature via the backbone, followed by three well-designed components to predict specific information. We leverage Part Perception Module to regress seven adaptive human-part related points from the assumed center for each human instance. Then we conduct the receptive field adaptation  in Enhanced Center-aware Branch by aggregating the features of the adaptive points to predict center heatmap. In addition, Two-hop Regression Branch adopts the adaptive human-part related points as one-hop nodes to indirectly regress the offsets from center to each keypoint. 

\begin{figure}
\begin{center}
\includegraphics[height=0.454\columnwidth,width=1\columnwidth, trim=65 85 65 85]{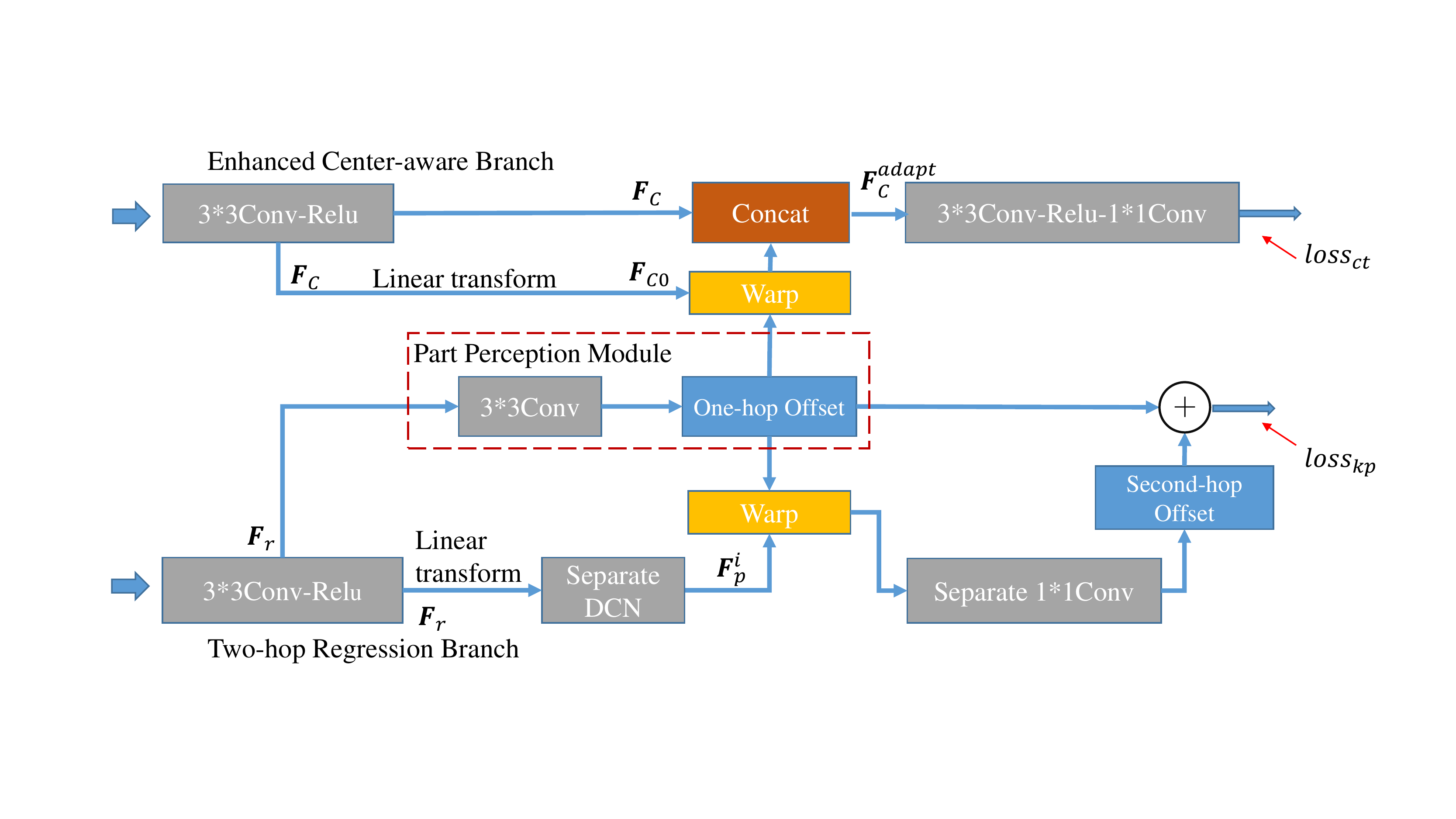}
\end{center}
\caption{The detailed structure across the Part Perception Module, Enhanced Center-aware Branch and Two-hop Regression Branch. Linear transform indicates the feature compression along the channel dimension via 1*1 convolution. The red arrows is used to indicate where a loss is applied.}
\label{fig:image4}
\end{figure}

{\noindent\bf Part Perception Module.} Based on the novel representation, we artificially divide each human instance into seven fine-grained parts (i.e. face, shoulder, left arm, right arm, hip, left leg, right leg) according to the inherent structure of human body. Part Perception Module is proposed to perceive the human parts by predicting seven adaptive human-part related points. For each part, we automatically regress an adaptive point to represent it without any explicit supervision. As shown in Figure \ref{fig:image4}, the feature $F_{k}$ is fed to the 3$\times$3 convolutional layer to regress 14-channel x-y offsets from the center to seven adaptive human-part related points. These adaptive points act as intermediate nodes, which are used for subsequent predictions.

{\noindent \bf Enhanced Center-aware Branch.} In previous works \cite{zhou2019objects,nie2019single}, the center of human instances with various scales and pose are predicted via the feature with fixed receptive field for each position. We propose a novel Enhanced Center-aware Branch to aggregate the features of seven adaptive human-part related points for precise center estimation. It can be considered as receptive field adaptation process as well as capture the variation of pose deformation and scales. 



As shown in Figure~\ref{fig:image4}, We use the structure of~3$\times$3 conv-relu to generate the branch-specific feature. In Enhanced Center-aware Branch, $\mathbf{F}_{c}$ is branch-specific feature with the fixed receptive field for each position. We firstly conduct linear transform that compressing the feature~$\mathbf{F}_{c}$ to obtain the squeezed feature~$\mathbf{F}_{c0}$ and then extract the feature vectors of the adaptive points via bilinear interpolation on $F_{c0}$, which can be deemed as a warp operation mentioned in Figure~\ref{fig:image3}. We named the extracted features as \{$\mathbf{F}_{c}^{face}$,~$\mathbf{F}_{c}^{sho}$,~$\mathbf{F}_{c}^{la}$,~$\mathbf{F}_{c}^{ra}$,~$\mathbf{F}_{c}^{hip}$,~$\mathbf{F}_{c}^{ll}$,~$\mathbf{F}_{c}^{rl}$\} corresponding to seven manually divided human parts (i.e., face, shoulder, left arm, right arm, hip, left leg, right leg) and concatenate them with $\mathbf{F}_{c}$ to generate the feature $\mathbf{F}_{c}^{adapt}$. Since the predicted adaptive points located on the seven divided parts are relatively evenly distributed on the human body region, thus the process above can be regarded as the receptive field adaptation according to the human scale as well as finely-grained capture the various body deformation. Finally we use $\mathbf{F}_{c}^{adapt}$ with adaptive receptive field to predict the 1-channel probability map for the center localization.

We use a Gaussian distribution $\mathbf{G}_{xy}= \exp{(-\frac{(x-{C}_{x})^{2}+(y-{C}_{y})^{2}}{2*{\delta}^{2}})}$ with mean $({C}_{x},{C}_{y})$ and adaptive variance $ \delta $ to generate the ground-truth center heatmap. For the loss function of the Enhanced Center-aware Branch, we employ the pixel-wise focal loss in a  penalty-reduced manner as follows:

\begin{equation} \label{formula:1}
{loss}_{ct}= \frac{1}{N} \sum_{n=1}^{N} 
\left\{
\begin{aligned}
(1-\bar{P}_{c})^{\alpha}\;\ln{(\bar{P}_{c})} & & if {P}_{c}=1 \\
(1-{P}_{c})^{\beta} \; \bar{P}_{c}^{\alpha} \; \ln(1-\bar{P}_{c}) & & elif {P}_{c} \not= 1  ,\\
\end{aligned}
\right.
\end{equation}
where N refers to the number of positive sample, $\bar{P}_{c}$ and ${P}_{c}$ indicate the predicted confidence scores and corresponding ground truth. $\alpha$ and $\beta$ are hyper-parameters and set to 2 and 4, following \cite{zhou2019objects}.

{\noindent\bf Two-hop Regression Branch.}  We leverage a two-hop regression method to predict the displacements instead of directly regressing the center-to-joint offsets. In this manner, the adaptive human-part related points predicted by Part Perception Module act as one-hop nodes to adaptively factorize long-range center-to-joint offsets into center-to-part and part-to-joint offsets.


Similar to Enhanced Center-aware Branch, we firstly leverage the structure of 3$\times$3 conv-relu to generate branch-specific feature, named $\mathbf{F}_{r}$ in Two-hop Regression Branch. Then we feed $\mathbf{F}_{r}$ into the separate deformable convolutions \cite{XizhouZhu2019DeformableCV,JifengDai2017DeformableCN} to generate separate features $\mathbf{F}_{p}^{i}$ for each human part. Then we extract the feature vectors at the adaptive points via a warp operation on corresponding feature $\mathbf{F}_{p}^{i}$. For convenience, we denote the extracted features as \{$\mathbf{F}_{face}$,~$\mathbf{F}_{sho}$, ~$\mathbf{F}_{la}$,~$\mathbf{F}_{ra}$,~$\mathbf{F}_{hip}$,~$\mathbf{F}_{ll}$,~$\mathbf{F}_{rl}$\} corresponding to seven divided parts (i.e., face, shoulder, left arm, right arm, hip, left leg and right leg). The extracted features are responsible for locating the keypoints belong to corresponding part by separate 1$\times$1 convolutional layers. For example, $\mathbf{F}_{face}$ is used to localize the eyes, ears and nose, $\mathbf{F}_{la}$ is used to localize the left wrist and left elbow, $\mathbf{F}_{ll}$ is used to localize the left knee and left ankle. 

Two-hop Regression Branch outputs a 34-channel tensor corresponding to x-y offsets $\bar{\mathbf{off}}$ from the center to 17 keypoints, which is regressed by a two-hop manner as follows:   

\begin{equation}
   \bar{\mathbf{off}} = \bar{\mathbf{off}_{1}}+\bar{\mathbf{off}_{2}},
\end{equation}
where $\bar{\mathbf{off}_{1}}$ and $\bar{\mathbf{off}_{2}}$ respectively indicate the offset from center to adaptive human-part related point~(one-hop offset mentioned in Figure~\ref{fig:image4}) and the offset from human-part related point to specific keypoints~(second-hop offset mentioned in Figure~\ref{fig:image4}). The predicted offsets $\bar{\mathbf{off}}$ are supervised by L1 loss and the supervision only acts at positive keypoint locations, the other negative locations are ignored. The loss function is formulated as follows:
\begin{equation}
   {loss}_{kp}= \frac{1}{K} \sum_{n=1}^{K} \left|  \bar{\mathbf{off}^{n}}- \mathbf{off}^{n}_{gt} \right|,
\label{formula:2}
\end{equation}
where $\mathbf{off}^{n}_{gt}$ is the ground truth offset from center to each joint. $K$ is the number of positive keypoint locations.

\begin{table*}
\begin{center}
\resizebox{1.8\columnwidth}{!}{

\begin{tabular}{l|c|ccccc|c}
\hline
Methods &Params & $AP$ & $AP_{50}$ & $AP_{75}$ & $AP_{M}$ & $AP_{L}$ &Time(s) \\  
\hline

\hline
\multicolumn{7}{c}{Bottom-up Heatmap-based Methods}\\
\hline
CMU-Pose$^{*\dag}$~\cite{cao2017realtime} &-&61.8& 84.9& 67.5& 57.1 &68.2& 0.077 \\

AE$^{*\dag}$~\cite{newell2017associative} &227.8 & 65.5 & 86.8 & 72.3 & 60.6 & 72.6&0.25  \\
CenterNet-DLA~\cite{zhou2019objects} &- &57.9&84.7&63.1&52.5&67.4&0.044\\
CenterNet-HG~\cite{zhou2019objects} &- & 63.0 & 86.8 & 69.6 & 58.9 & 70.4&0.152 \\
PersonLab~\cite{papandreou2018personlab} &68.7 & 66.5& 88.0& 72.6& 62.4& 72.3&- \\
PifPaf~\cite{kreiss2019pifpaf} & - & 66.7 & - & -& 62.4& 72.9&- \\
HrHRNet-W48$^{*\dag}$~\cite{cheng2020higherhrnet} & 63.8& 70.5 & 89.3 & 77.2 & 66.6&75.8 & 0.182  \\
FCPose (ResNet-101) \cite{mao2021fcpose} & -&65.6&87.9&72.6&62.1& 72.3 & 0.093\\
SWAHR-W48$^{*}$ \cite{luo2021rethinking} &63.8&70.2&89.9&76.9&65.2&77.0& 0.16\\

\hline
\multicolumn{7}{c}{Single-stage Regression-based Methods}\\
\hline
SPM $^{* \dag}$~\cite{nie2019single} & - & 66.9& 88.5& 72.9& 62.6& 73.1&0.058 \\
DirectPose $^{\dag}$~\cite{tian2019directpose} &- &64.8& 87.8& 71.1& 60.4& 71.5&0.087 \\
PointSetNet $^{* \dag}$~\cite{wei2020point} &- &68.7& 89.9& 76.3& 64.8& 75.3& - \\
DEKR-W32$^{* \dag}$~\cite{geng2021bottom} & 29.6 &69.8& 89.0 & 76.6 & 65.2 & 76.5 &0.192 \\
DEKR-W48$^{* \dag}$~\cite{geng2021bottom} & 65.7 & 71.0 & 89.2 & 78.0 & 67.1& 76.9 & 0.224 \\
\hline

Ours~(DLA-34)$^{\dag}$ &21.0& 67.4 & 88.2 & 73.7 & 63.2 & 74.7&{\bf0.034}  \\
Ours~(DLA-34+)$^{\dag}$ &21.0& 68.2 & 89.0 & 75.1 & 64.6 & 75.0&0.045   \\
\bf{Ours~(HRNet-W48)}$^{\dag}$ & 64.7 & {\bf71.3} & {\bf90.0} & {\bf78.3} & {\bf67.1} & {\bf77.2}&0.110    \\
\hline
\end{tabular}
}
{\caption{Comparisons with previous state-of-the-art methods on COCO test-dev set. ${*}$ indicates using extra test-time refinements. $\dag$ refers to multi-scale testing. DLA-34+ indicates DLA-34 with 640 pixels input resolution. Note that the reported inference time of HigherHRNet \cite{cheng2020higherhrnet} and SWAHR \cite{luo2021rethinking} exclude the test refinement time.}\label{tab:test}}
\end{center}
\end{table*}

\subsection{Training and Inference Details}\label{subsection3.3}



During training, we employ an auxiliary training objective to learn keypoint heatmap representation, which enables the feature to maintain more human structural geometric information. In particular, we add a parallel branch to output a 17-channel heatmap corresponding to 17 keypoints and apply a Gaussian kernel with adaptive variance to generate ground truth keypoint heatmap. We denote this training objective as ${loss}_{hm}$, which is similar to Equation \ref{formula:1}. The only difference is that N refers to the number of positive keypoints. The auxiliary branch is only used for training process and removed in inference process. Our total training target function for multi-task training process is formulated as:
\begin{equation}
   {loss}_{total}= {loss}_{ct} + {loss}_{kp} + {loss}_{hm}.
\end{equation}

During inference, Enhanced Center-aware Branch outputs the center heatmap that indicates whether the position is center or not. Two-hop Regression Branch outputs the offsets from the center to each joint. We firstly pick the human center by using 5$\times$5 max-pooling kernel on the center heatmap to maintain 20 candidates, and then retrieve the corresponding offsets $(\delta_{x}^{i},\delta_{y}^{i})$ to form human pose without any post-process and extra refinement. Specifically, we denote the predicted center as $({C}_{x},{C}_{y})$. The above decode process is formulated as follows:

\begin{equation}
   ({K}_{x}^{i},{K}_{y}^{i})= ({C}_{x},{C}_{y}) + (\delta_{x}^{i},\delta_{y}^{i}) ,
\end{equation}
where $({K}_{x}^{i},{K}_{y}^{i})$ is the coordinate of the i-th keypoint. In contrast to DEKR\cite{geng2021bottom} that uses the average of the heat values at each regressed keypoints via bilinear interpolation, we only leverage the center heat-values as the pose score for fast inference.


\section{Experiments and Analysis}

In this section, we first briefly introduce the dataset, evaluation metric, data augmentation and implementation details. Next, we compare our proposed method with the previous state-of-the-art methods. Finally,  we conduct comprehensive ablation study to reveal the effectiveness of each component.

\subsection{Experimental Setup}\label{subsection4.1}

\noindent{\bf Dataset.} The COCO dataset~\cite{lin2014microsoft} consists of over 200,000 images and 250,000 human instances labeled with 17 keypoints for pose estimation task. It is divided into train, mini-val, test-dev sets respectively. We train our model on COCO train2017 dataset. The comprehensive experimental results are reported on the COCO mini-val set with 5000 images and test-dev2017 set with 20K images.


\begin{table}
\begin{center}

\resizebox{1\columnwidth}{!}{
\begin{tabular}{l|cccc|ccccc}
\hline
$ Expt.$ & $PPM$ & $RFA$ & $TR$ & $AL$ & $AP$ & $AP_{50}$ & $AP_{75}$ & $AP_{M}$ & $AP_{L}$  \\
\hline
$1$&- & -&- & -&  56.0 & 82.7 & 60.0 & 47.3 & 68.4 \\
$2$&$\surd$ & $\surd$ & -& -& 57.3 & 83.0 & 61.8 & 48.7 & 69.1 \\
$3$&$\surd$ & - &$\surd$ & -& 60.0 & 84.3 & 65.0 & 51.7 & 71.0  \\                        
$4$ & $\surd$ & $\surd$ &$\surd$ &- & 61.6& 85.2&67.5&53.7&72.8 \\


$5$&$\surd$ & $\surd$ &$\surd$ & $\surd$ & {\bf64.9} & {\bf86.4} & {\bf70.9} & {\bf58.6} & {\bf74.2}  \\
\hline

\end{tabular}}
{\caption{Ablation studies. $PPM$ denotes Part Perception Module, $RFA$ is receptive field adaptation conducted in Enhanced Center-aware Branch, $TR$ indicates two-hop regression strategy in Two-hop Regression Branch, $AL$ refers to employ an auxiliary loss ${loss}_{hm}$ to learn keypoint heatmap representation.}\label{tab:c}}
\end{center}
\vspace{-1mm}
\end{table}

\noindent{\bf Evaluation Metric.} We leverage average precision and average recall based on Object Keypoint Similarity~(OKS) to evaluate our keypoint detection performance. 

\noindent{\bf Data Augmentation.} During training, we use random flip, random rotation, random scaling and color jitter to augment training samples. The flip probability is 0.5, the rotation range is (-30, 30) and the scale range is (0.6, 1.3). Each input image is cropped according to the random center and random scale then resized to 512 / 640 pixels for DLA-34~\cite{yu2018deep} and 800 pixels for HRNet-W48~\cite{sun2019deep}. The output size is 1/4 of the input resolution.

\noindent{\bf Implementation Details.} We train our proposed model via Adam optimizer with a mini-batch size of 64 (8 per GPU) on a workstation with eight 12GB Titan Xp GPUs. We use initial learning rate of 2.5e-4. All codes are implemented with Pytorch. DLA-34~(19.7M) and HRNet-W48~(63.6M) are adopted to achieve the trade-off between the accuracy and efficiency. For inference, we keep the aspect ratio and  resize the short side of the images to 512 / 640 / 800 pixels. The inference time is calculated on a 2080Ti GPU with mini-batch 1. We further use flip and multi-scale image pyramids to boost the performance. It is worth highlighting that the flip operation is only applied to the center heatmap predicted by Enhanced Center-aware Branch.  


\subsection{Comparison with the State-of-the-art Methods}\label{subsection4.2}

\noindent{\textbf{Test-dev Results.}}~ As reported in Table~\ref{tab:test}, we compare our method with the previous state-of-the-art methods. In details, for bottom-up methods, our method achieves 67.4 AP, which outperforms the widely-used CMU-Pose~\cite{cao2017realtime}, AE~\cite{newell2017associative} as well as CenterNet-HG \cite{zhou2019objects} on both performance and speed with a smaller model DLA-34. In contrast to HigherHRNet-W48 \cite{cheng2020higherhrnet} that uses higher resolution heatmap and test-time refinement to generate the final results, our approach with HRNet-W48 achieves 0.8 AP improvements without extra upsample and refinement operation.  

For single-stage regression-based methods, our method with HRNet-W48 surpasses SPM~\cite{nie2019single}~(refined by the well-trained single-person pose estimation model) by 4.4\% AP without any refinement and also outperforms DirectPose~\cite{tian2019directpose} with a large margin by 6.5\% AP. In comparison to state-of-the-art DEKR \cite{geng2021bottom}, we achieve 0.3 AP gain without extra pose NMS  and pose scoring network during inference. The comprehensive comparisons prove that our method achieves the better performance with the more efficient and compact pipeline.

\subsection{Ablative Analysis}\label{subsection4.3}
All ablation studies adopt DLA-34 as backbone and use the 1x training epoch (140 epochs) with single-scale testing on the COCO mini-val set. 


\begin{table}
\begin{center}

\resizebox{0.8\columnwidth}{!}{
\begin{tabular}{l|ccccccc}
\hline
$shared$ & $AP$ & $AP_{50}$ & $AP_{75}$ & $AP_{M}$ & $AP_{L}$ &  $AR$  \\
\hline
 
 - & 64.2 & 86.0 & 70.3 & 57.9 & 73.6 & 69.9 \\
 $\surd$ &{\bf64.9} & {\bf86.4} & {\bf70.9} & {\bf58.6} & {\bf74.2} & {\bf70.5} \\
\hline

\end{tabular}}
{\caption{$Shared$ refers to whether the adaptive points is shared between Enhanced Center-aware Branch and Two-hop Regression Branch.}\label{tab:1}}
\end{center}
\vspace{-1mm}
\end{table}

\begin{figure*}
\begin{center}
\includegraphics[height=0.651\columnwidth,width=2.0\columnwidth]{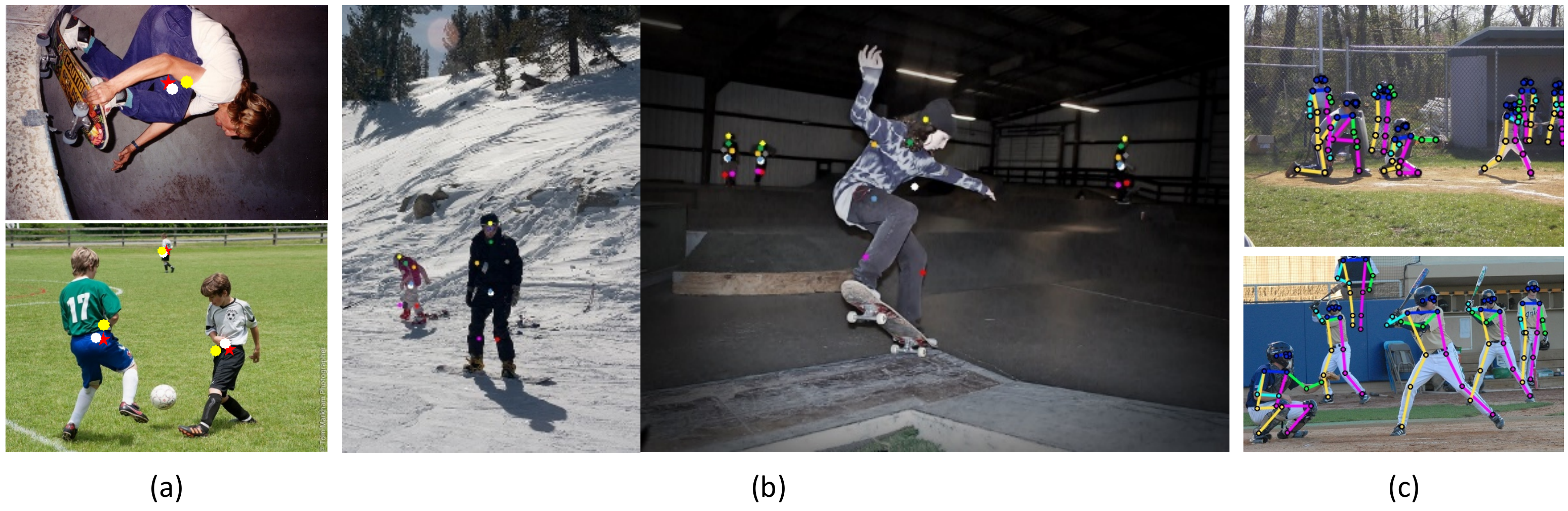}
\end{center}
\caption{(a) Visualization of ground truth center (red star), the predicted center with receptive field adaptation (white point) and the predicted center without receptive field adaptation (yellow point).  (b) Visualization of the predicted adaptive point set including the center and seven human-part related points for each human instance on COCO dataset. (b) Examples of predicted skeleton with various pose deformation on COCO dataset.}
\label{fig:image7}
\vspace{-1mm}
\end{figure*}

\noindent{\bf Analysis of Part Perception Module.} We conduct the controlled experiments to study the usage of Part Perception Module.~As reported in Table~\ref{tab:1}, we respectively conduct the experiments that using shared adaptive points and unshared adaptive points (automatically locating seven positions via an additional 3$\times$3 convolution) between Enhanced Center-aware Branch and Two-hop Regression Branch. Using the shared adaptive points obtains 0.7\% AP improvements with less parameters and FLOPs compared with using unshared adaptive points~(64.9\% versus 64.2\%). We consider that using the shared adaptive points is more interpretable since Part Perception Module is simultaneously supervised by ${loss}_{ct}$ and ${loss}_{kp}$ that mentioned in Equation \ref{formula:1} and \ref{formula:2}. ${loss}_{ct}$ enables the adaptive points to more concentrate on the region with semantic information. ${loss}_{kp}$ drives the adaptive points to perceive the divided human parts so as to better conduct receptive field adaptation in Enhanced Center-aware Branch. 

\noindent{\bf Analysis of Enhanced Center-aware Branch.} We conduct the controlled experiments to explore the effect of receptive field adaptation process in Enhanced Center-aware Branch. Compared with using the feature with fixed receptive field to perceive the human center, receptive field adaptation process obtains 1.6\% AP improvements in ($Expt.$~3 versus $Expt.$~4) of Table~\ref{tab:c}. The results prove that receptive field adaptation is capable of finely-grained capturing body deformation and dynamically adjusting the receptive field for the center of human instances with various scales, thus it is able to perceive the instance center more precisely. As shown in Figure \ref{fig:image7}(a), with receptive field adaptation, the predicted center is more closer to ground-truth center than without receptive field adaptation applied.

\begin{table}
\begin{center}

\resizebox{0.9\columnwidth}{!}{
\begin{tabular}{l|cccccc}
\hline
 Methods &  $AP$ & $AP_{50}$ & $AP_{75}$ & $AP_{M}$ & $AP_{L}$ &  $AR$    \\
\hline

$HKR$ & 58.5 & 83.3 & 63.8 & 50.0 & 69.4 & 66.8 \\
$TR$ &{\bf60.0}& {\bf84.3} & {\bf65.0} & {\bf51.7}& {\bf71.0}& {\bf 68.0} \\
\hline
\end{tabular}
}
{\caption{Comparisons between the hierarchical keypoint regression (notated as $HKR$) in SPM and our adaptive Two-hop Regression (notated as $TR$).}\label{tab:error}}
\end{center}
\vspace{-1mm}
\end{table}

\noindent{\bf Analysis of Two-hop Regression Branch.} We conduct the controlled experiments ($Expt.$~1 versus $Expt.$~3) in Table~\ref{tab:c} to investigate the effect of two-hop regression. It brings to 4.0\% AP improvements compared with directly regressing the displacements from center to each joints. Furthermore, based on $Expt.$~1, we implement the hierarchical body representation used in SPM to hierarchically regress the keypoint. As shown in Table \ref{tab:error}. It drops 1.5 AP compared with our Two-hop Regression. The results prove our two-hop regression approach is capable of factorizing long-range center-to-joint offsets and avoiding the accumulated errors.


\noindent{\bf Analysis of auxiliary loss.} ($Expt.$~4 versus $Expt.$~5) studies the effect of auxiliary loss, we achieve improvements of 3.3\% AP by employing auxiliary loss to help training. It experimentally proves that the keypoint heatmap is able to retain more structural geometric information to improve regression performance.


\noindent{\bf Analysis of Heatmap Refinement for our regression result.} For validating our regression performance, we further snap the closest confidence peaks on the keypoint heatmap to refine the regressed predictions. For convenience, the two-hop regressed result and the two-hop regressed result with heatmap refinement are denoted as $Ours\_reg$ and $Ours\_heat$ respectively. As shown in Table~\ref{tab:5}, $Ours\_reg$ obtain slightly better performance than $Ours\_heat$~(64.9\% AP versus 64.6\% AP), which proves the heatmap refinement is negligible for our two-hop regression method.

\noindent{\textbf{Qualitative Results.}} We visualize some qualitative results generated by our AdaptivePose. Figure~\ref{fig:image7}(b) shows the predicted human center and seven adaptive human-part related points conditioned on each human instance with various scales and pose. The visualizations prove that the predicted adaptive human-part related points are able to finely-grained capture the human parts with various deformation and scales. In Figure~\ref{fig:image7}(c), there are examples of the predicted skeleton on COCO mini-val set, which contain the diverse human bodies with various deformation.

\begin{table}
\begin{center}

\resizebox{0.9\columnwidth}{!}{
\begin{tabular}{l|cccccc}
\hline
 Methods &  $AP$ & $AP_{50}$ & $AP_{75}$ & $AP_{M}$ & $AP_{L}$ &  $AR$    \\
\hline

$Ours\_heat$ & 64.6 & {\bf87.0} & 70.3 & 58.0 & 74.1 & 69.7 \\
$Ours\_reg$ &{\bf64.9}& 86.4 & {\bf70.9} & {\bf58.6}& {\bf74.2}& {\bf 70.5} \\
\hline
\end{tabular}
}
{\caption{Ablation study for exploring the effect of heatmap refinement for our two-hop regressed result. }\label{tab:5}}
\end{center}
\vspace{-1mm}
\end{table}


\section{Conclusion}

In this paper, we propose to represent the human parts as points and introduce an adaptive body representation, which represents human body in a fine-grained fashion. Based on the proposed body representation, we construct a single-stage network, termed AdaptivePose, which including three effective components: Part Perception Module, Enhanced Center-aware Branch and Two-hop Regression Branch. During inference, we eliminate the grouping as well as refinements, and only need a single-step process to form human pose. We experimentally prove that AdaptivePose obtains the best speed-accuracy trade-off and outperforms previous state-of-the-art bottom-up and single-stage methods.

\section{Acknowledgements}
This work is supported by the National Natural Science Foundation of China (62071056) the Doctorial Innovation Foundation of Beijing University of Posts and Telecommunications (CX2020111).




\end{document}